\documentclass{article}
\usepackage{spconf,amsmath,graphicx}
\usepackage{kotex}
\usepackage{graphicx}
\usepackage{subcaption}
\usepackage{wrapfig}
\usepackage{amsmath}
\usepackage{multirow}
\usepackage{balance}
\usepackage{enumitem}

\title{Speech Tokenizer is Key to Consistent Representation}
\name{Wonjin Jung, Sungil Kang, Dong-Yeon Cho}
\address{SK Telecom}
\begin{document}
%\ninept
%
\maketitle
% ===========================================================================================================================
\begin{abstract}
Speech tokenization is crucial in digital speech processing, converting continuous speech signals into discrete units for various computational tasks. This paper introduces a novel speech tokenizer with broad applicability across downstream tasks. While recent advances in residual vector quantization (RVQ) have incorporated semantic elements, they often neglect critical acoustic features. We propose an advanced approach that simultaneously encodes both linguistic and acoustic information, preserving prosodic and emotional content. Our method significantly enhances speech representation fidelity across diverse applications. Empirical evaluations demonstrate its effectiveness in speech coding, voice conversion, emotion recognition, and multimodal language modeling, without requiring additional training. This versatility underscores its potential as a key tool for advancing AI-driven speech processing.
\end{abstract}
\begin{keywords}
Speech quantization, speech coding, speech LLM
\end{keywords}
% ===========================================================================================================================
% ===========================================================================================================================
\section{Introduction}
\label{sec:intro}
The human brain's remarkable ability to process diverse speech inputs raises a compelling question: Can we create AI systems that match this versatility, tackling multiple speech tasks without specific training?
% The human brain's auditory system shows remarkable flexibility in processing various speech inputs, demonstrating task-agnostic problem-solving abilities across a wide range of speech-related challenges. This mental adaptability raises a key question in the field of artificial intelligence: Can we create speech processing systems that mimic this human-like versatility, able to generalize across multiple downstream tasks without task-specific training? 

Recent progress in Large Language Model (LLM) has brought us closer to achieving this goal, demonstrating exceptional performance in various speech understanding and generation tasks. Central to this paradigm shift is the crucial role of effective speech tokenization~\cite{tradSpeechCodec}, serving as the foundation for connecting raw speech signals to the symbolic representations~\cite{Santoso2024}.
A key challenge in applying speech signals is efficiently representing both semantic and acoustic information. SoundStream introduced an efficient method using residual vector quantization (RVQ) for embedding features~\cite{SoundStream2021}. EnCodec incorporated Long Short-Term Memory (LSTM) layers to improve sequential signal modeling~\cite{Encodec2022}. These neural speech codecs typically operate in three stages: encoding, quantization, and decoding~\cite{kim2024}.
Improvements focus on enhancing quantization efficiency and embedding features. For efficient quantization, methods using adversarial and reconstruction loss have been explored~\cite{singlecodec}. WavTokenizer extended the codebook space in VQ~\cite{WavTokenizer2024}. Approaches considering inter-frame correlations have been proposed to capture temporal dependencies~\cite{interframe2024, JUNG2013}.
For embedding improvements, RepCodec focuses on semantic token generation~\cite{RepCodec}, while SpeechTokenizer aligns embedding features with semantic information~\cite{SpeechTokenizer2024}. Training of discrete tokens with LLMs has achieved effective multimodal performance~\cite{anygpt2024}. Incorporating semantic information improves compression efficiency and enhances speech understanding when integrated with LLMs.
However, as discrete token applications expand, preserving acoustic information has become more important, exposing limitations of existing methods. 
In this paper, we propose a novel neural codec framework that simultaneously learns both acoustic and semantic information. Our approach leverages the HuBERT model~\cite{HuBERT2021} for semantic feature extraction and the ECAPA model~\cite{Ecapa2020} for robust handling of acoustic features. The quantized discrete tokens can be directly applied to autoregressive models (\textit{e.g.}, LLM), enabling seamless integration into various downstream tasks without task-specific fine-tuning.
The distinguishing feature of our proposed method, compared to the existing SpeechTokenizer, our method incorporates not only the semantic information of speech but also acoustic information as prior knowledge. This approach enables the model to learn a more comprehensive representation of speech components. This enhanced representation of speech components demonstrates the efficacy of our proposed approach across 4 downstream tasks, where it outperforms the performance of existing methods.
% To validate the effectiveness of our approach, we observed about a 0.5-point improvement in Perceptual Evaluation of Speech Quality (PESQ) scores compared to existing frameworks~\cite{PESQ}. To confirm that acoustic information is effectively captured in the discrete tokens, we applied our method to an SER task, recognizing 4 emotion classes. Our model improved accuracy by 10\%, capturing emotions that previous methods could not effectively express. In the ASR task, our approach maintained a similar word error rate (WER) to prior methods for semantic information representation. Additionally, by swapping acoustic tokens between speakers without further training, we successfully demonstrated effective voice conversion, showcasing the disentanglement of semantic and acoustic information in the representation.

Our main contributions are as follows:
\begin{enumerate}[label=--]
    \item \textbf{Unified acoustic and semantic information framework}: We propose a novel neural speech codec that jointly learns acoustic and semantic information by integrating robust acoustic and semantic features.
    \item \textbf{Improved speech coding efficiency and broad downstream task applicability}: Our approach improves perceptual performance over existing frameworks, enabling clearer speech reconstruction. It also effectively represents speech information across various downstream tasks.
    \item \textbf{Versatile speech token}: The quantized discrete tokens can be directly utilized in LLMs, demonstrating broad applicability for text token-based LLM training.
\end{enumerate}

% \begin{figure*}[t!]
%     \centering
%     % \includegraphics[width=0.9\textwidth]{ver2_fig/fig2.pdf}
%     % \includegraphics[width=0.9\textwidth]{ver4_fig/FIG1.pdf}
%     \includegraphics[width=0.9\textwidth]{ver4_fig/fig1_4.pdf}
%     \centering
%     \caption{Proposed framework}
%     \label{fig2_architecture:proposed_framework}
% \end{figure*}
\section{Method}
\label{sec:format}
Our proposed method focuses on designing an efficient speech tokenization system that jointly considers semantic and acoustic information. 
We employ Encodec to embed speech and apply RVQ for efficient coding.
In this process, the initial tokenization captures semantic information related to linguistic meaning. Subsequently, the residual values, after subtracting the linguistic information, learn to capture acoustic information, allowing them to represent acoustic features.
The quantized speech information encapsulates the most salient features, including semantic and acoustic data, while the residuals retain additional information from the speech signal.
In this section, we discuss the learning methodologies optimized to represent each type of information accurately and explain the overall loss function used for reconstructing the speech signal.

\subsection{Basic model}
Our proposed method builds upon the Encodec architecture~\cite{Encodec2022}, an efficient speech codec inspired by Soundstream~\cite{SoundStream2021}. Encodec excels in compressing and reconstructing speech signals through a sophisticated encoding-decoding process.
The Encodec structure operates by encoding input speech signals into a latent space, followed by RVQ for further compression. This process can be formalized as:
\begin{equation}
    \mathbf{z}_0 = E(\mathbf{x})
\end{equation}
where $\mathbf{x}$ is the input speech signal and $E(\cdot)$ is the encoder function that maps the input to its initial embedding feature $\mathbf{z}_0$. 
The embedding feature is then processed by RVQ for further compression. Each residual representation $\mathbf{z}_{i}$ captures additional information from the signal:
\begin{equation}
    \mathbf{z}_{i+1} = \mathbf{z}_i - \mathbf{\hat{z}}_i, \quad \mathbf{\hat{z}}_i = \text{VQ}(\mathbf{z}_i)
\end{equation}
where $\text{VQ}(\cdot)$ denotes the vector quantization operation.
The decoding stage reverses this process, reconstructing the speech signal from the quantized representations:
\begin{equation}
    \mathbf{\hat{x}} = D(\mathbf{\hat{z}}_0, \mathbf{\hat{z}}_1, ..., \mathbf{\hat{z}}_N)
\end{equation}
where $D(\cdot)$ is the decoder function and $N$ is the number of quantization levels.

This approach enables Encodec to achieve high compression ratios while maintaining high-quality speech reproduction. Our method leverages this architecture to capture both semantic and acoustic information efficiently, improving performance in various speech processing tasks.
% ===========================================================================================================================

\subsection{Feature representation}
Speech signals contain both semantic and acoustic information. 
We design the first quantized value $\hat{\mathbf{z}}_{0}$ to represent semantic features, while its residual $\mathbf{z}_1$, quantized to obtain $\hat{\mathbf{z}}_1$, captures acoustic features. 
% Fig.\ref{fig2_architecture:proposed_framework} illustrates the detailed learning process.

% \subsubsection{Semantic Representation}
To represent semantic information, we extract feature vectors using HuBERT~\cite{HuBERT2021}. We align these feature vectors with the quantized $\hat{\mathbf{z}}_0$ values by learning the similarity between the two feature sets using cosine similarity loss:
\begin{equation}
% L_\text{semantic} = -\mathbb{E}\left[\log\left(\sigma\left(\cos\left(\text{feature}_{:,:n}, \text{target\_feature}_{:,:n}\right)\right)\right)\right]
\mathcal{L}_\text{semantic} = -\frac{1}{P} \sum_{d=1}^P \log \sigma(\cos(\mathbf{\hat{z}}_0, S(x)))
\end{equation}
where $S(x)$ is the output of HuBERT, $\mathbf{\hat{z}}_0$ is the quantized output, and $P$ represents the dimension of the semantic feature representation. This approach aligns with the methodology employed in SpeechTokenizer~\cite{SpeechTokenizer2024}.

% \subsubsection{Acoustic Representation}
The proposed method in this paper enhances the existing model by incorporating additional knowledge into the acoustic representation, resulting in a more effective learning approach.
The residual $\mathbf{z}_1$ obtained from the quantized $\hat{\mathbf{z}}_0$ focuses on acoustic information. We employ the ECAPA-TDNN model~\cite{Ecapa2020} to train the residual values to effectively represent acoustic information. To align the dimensions of the quantized residuals and ECAPA-TDNN's latent space, we introduce an additional align layer. This align layer employs a multi-head attention mechanism to effectively represent timbre information. To capture sequence timbre and prosody information more effectively, we implemented an attention mechanism with 8 multi-heads, enabling the model to learn richer and more diverse interactions within the speech signal. Additionally, the layer utilizes Kullback-Leibler divergence loss for matching to the same latent space:
\begin{equation}
% \mathcal{L}_\text{acoustic} = \sum \text{VQ}(\mathbf{r}_0) \log \frac{\text{VQ}(\mathbf{r}_0)}{A(x)}
\mathcal{L}_\text{acoustic} = \sum \hat{z}_1 \log \frac{\hat{z}_1}{A(x)}
\end{equation}
where $\hat{z}_1$ represents the quantization residual $\mathbf{z}_1$, and $A(x)$ denotes the continuous probability distribution generated by the align layer.

Through this learning process, each quantized embedding feature encapsulates both semantic and acoustic information present in the speech signal, represented by $S(x)$ and $A(x)$ respectively, providing a comprehensive representation for subsequent processing tasks.

\subsection{Training}
Our training approach enhances basic speech reconstruction by incorporating both semantic and acoustic information, where $x$ represents a speech signal and $\hat{x}$ denotes the reconstructed signal by the network. In the speech reconstruction process, we train the model with an additional discriminator loss term from the generative adversarial network training process.
\begin{equation}
\mathcal{L}_\text{Discriminator} = \text{Discriminator}(x, \hat{x})
\label{eq:Discriminator}
\end{equation}
where $\mathcal{L}_\text{Discriminator}$ is the discriminator loss, and Discriminator($x$, $\hat{x}$) represents the discriminator function comparing the original and reconstructed signals.
The generator loss is augmented by incorporating semantic and acoustic representation losses into the efficiently designed Encodec loss, which aims at reconstructing the original signal.
\begin{equation}
\begin{split}
% \mathcal{L}_\text{Generator} = &\ \mathcal{L}_\text{Encodec} \\
% & + \lambda_\text{semantic} \cdot \mathcal{L}_\text{semantic} + \lambda_\text{acoustic} \cdot \mathcal{L}_\text{acoustic}
\mathcal{L}_\text{Generator} = &\ \mathcal{L}_\text{Encodec} + \lambda_\text{s} \cdot \mathcal{L}_\text{semantic} + \lambda_\text{a} \cdot \mathcal{L}_\text{acoustic}
\label{eq:Generator}
\end{split}
\end{equation}
where $\mathcal{L}_\text{Generator}$ denotes the generator loss, and $\lambda_\text{s}$ and $\lambda_\text{a}$ represent the weighting factors between the respective features.

The discriminator is utilized to improve the quality of the reconstructed signal $\hat{x}$. The generator mainly utilizes the loss from $\text{Encodec}$, with additional loss terms introduced to encapsulate semantic and acoustic features in the encoder's output. To balance the overall loss, weighting factors $\lambda$ were empirically set to 0.5 for both factors during the training process.

\begin{figure}[t]
    \centering
    \begin{subfigure}[b]{0.48\columnwidth}
        \centering
        \includegraphics[width=\textwidth]{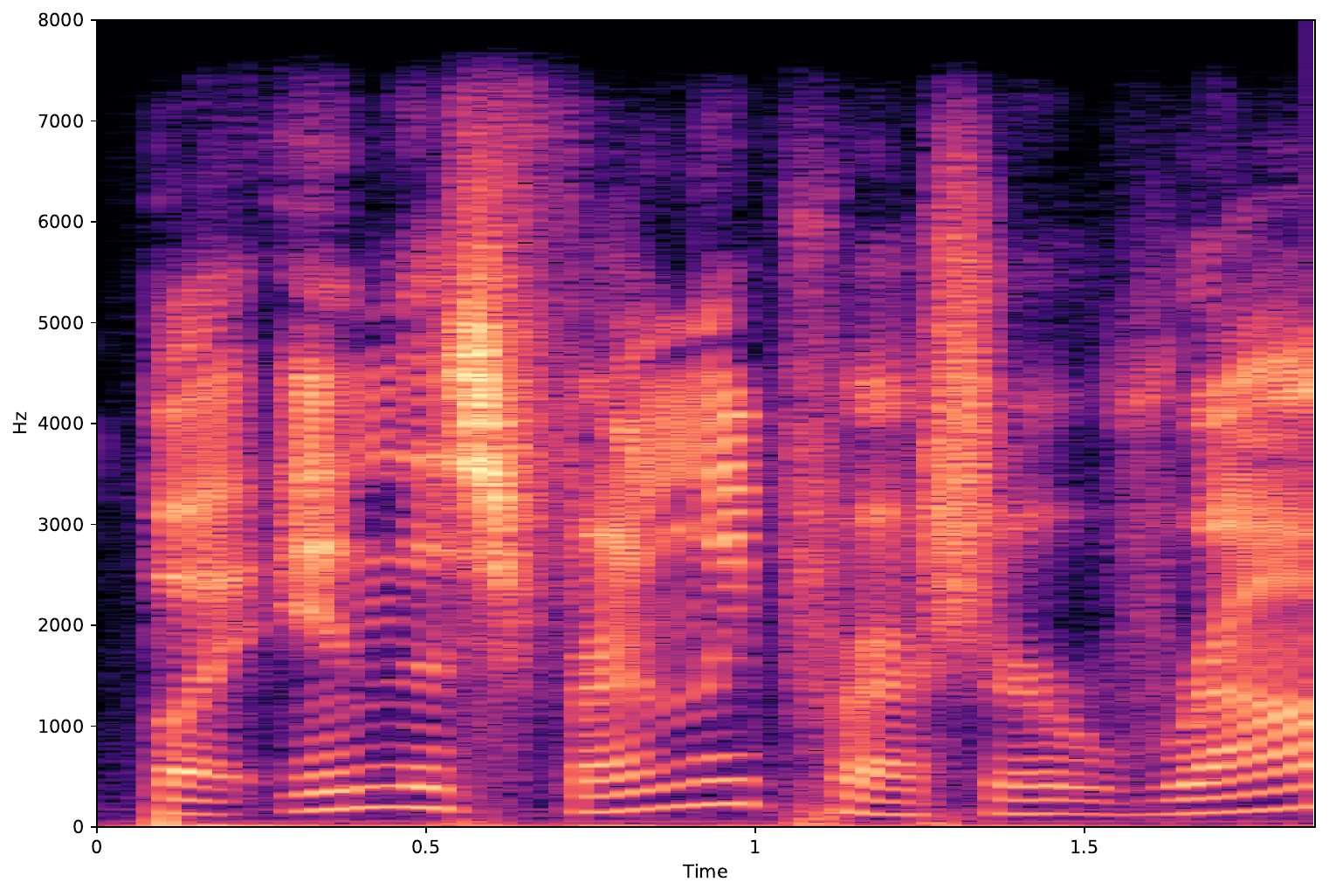}
        % \caption{Original speech spectrogram}
        \caption{}
        \label{fig:subfig1}
    \end{subfigure}
    \hfill
    \begin{subfigure}[b]{0.48\columnwidth}
        \centering
        \includegraphics[width=\textwidth]{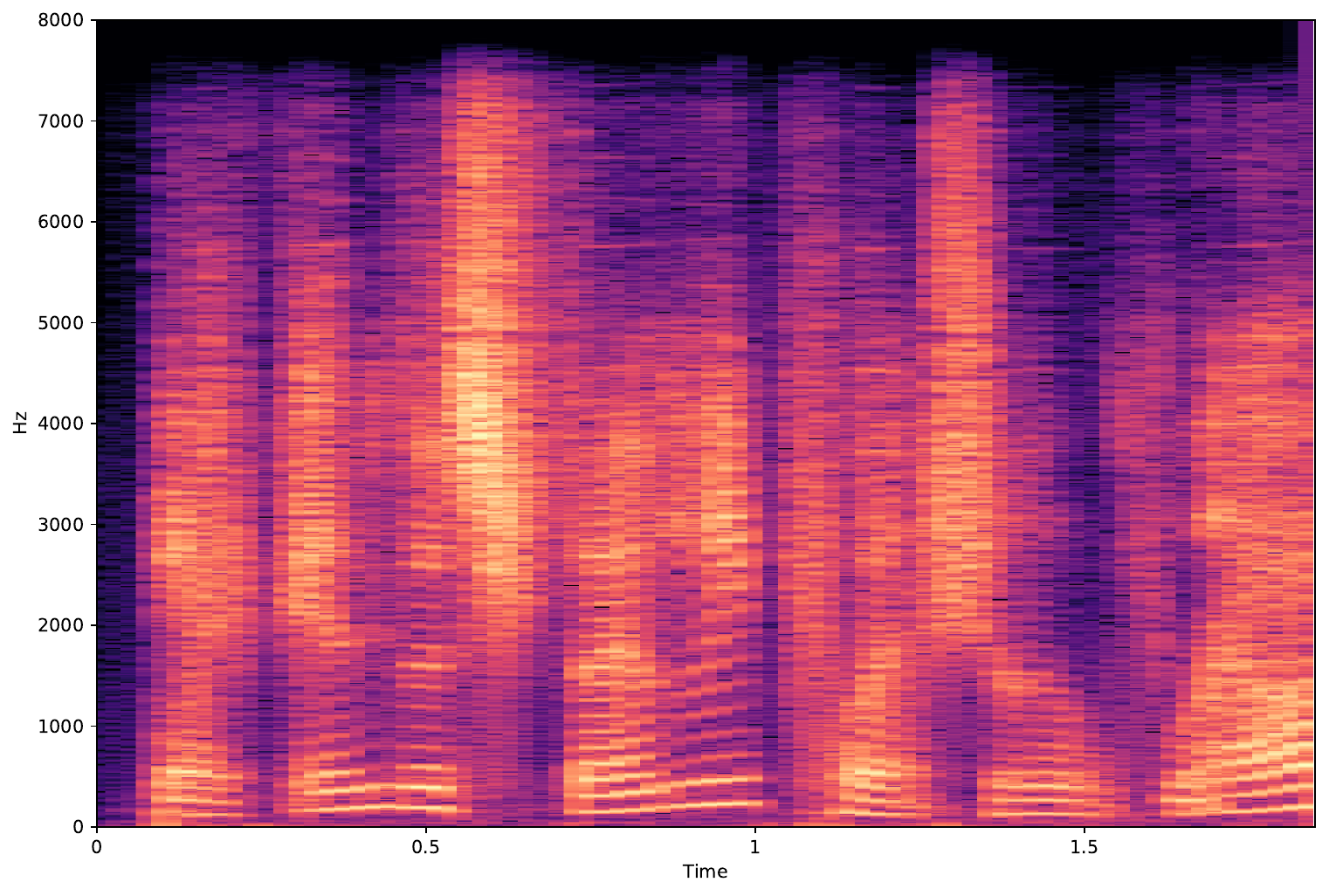}
        % \caption{reconstruction spectrogram by \textbf{our model}}
        \caption{}
        \label{fig:subfig3}
    \end{subfigure}
    \caption{Comparative analysis of speech spectrograms: (a) Spectrogram of the original signal; (b) Spectrogram reconstructed by our model using only 2 tokens per frame.}
    \label{fig:threefigures}
\end{figure}

\section{Experiments and Results}
\label{sec:typestyle}
For training our tokenizer model, we utilized the 16 kHz LibriSpeech 500-hour dataset~\cite{Librispeech}. Each speech frame was quantized at a 50 Hz frame rate (\textit{i.e.}, one frame per 20 ms). The model was trained for 10 epochs using a single A100 GPU. Encodec, SpeechTokenizer, and our proposed method were all trained under identical conditions. The learning rate was set to $1 \times 10^{-4}$, and the AdamW optimizer was employed with $\beta_{1}=0.9$ and $\beta_{2}=0.99$ to effectively apply weight decay. A VQ codebook of 1024 entries was used, and up to 8 residual tokens were employed.
Using the trained model, we conducted experiments on four downstream tasks without task-specific learning. These tasks include:
1) a speech codec for reconstruction, 2) voice conversion to demonstrate acoustic and semantic capabilities, 3) token-based Speech Emotion Recognition (SER) to verify acoustic information, and 4) token-based Automatic Speech Recognition (ASR) using LLM.
% 1) a speech codec for the reconstruction of the proposed model, 2) voice conversion to demonstrate the effectiveness of acoustic and semantic information between tokens, 3) token-based SER (Speech Emotion Recognition) to verify acoustic information, and 4) a speech token-based LLM for ASR (Automatic Speech Recognition) in multimodal LLMs.

\subsection{Speech Codec}
% \begin{figure*}[t]
%     \centering
%     \begin{subfigure}[b]{0.33\textwidth}
%         \centering
%         % \includegraphics[width=\textwidth]{ver4_fig/compare_spec_original_1.pdf}
%         \includegraphics[width=\textwidth]{ver4_fig/compare_spec_original.pdf}
%         \caption{Original speech spectrogram}
%         \label{fig:subfig1}
%     \end{subfigure}
%     \hfill
%     % \hspace{0.07\textwidth}
%     \begin{subfigure}[b]{0.33\textwidth}
%         \centering
%         % \includegraphics[width=\textwidth]{ver4_fig/compare_spec_default_1.pdf}
%         \includegraphics[width=\textwidth]{ver4_fig/compare_spec_default.pdf}
%         \caption{Speech spectrogram by SpeechTokenizer}
%         \label{fig:subfig2}
%     \end{subfigure}
%     \hfill
%     % \hspace{0.07\textwidth}
%     \begin{subfigure}[b]{0.33\textwidth}
%         \centering
%         % \includegraphics[width=\textwidth]{ver4_fig/compare_spec_proposed_1.pdf}
%         \includegraphics[width=\textwidth]{ver4_fig/compare_spec_proposed.pdf}
%         \caption{Speech spectrogram by \textbf{our model}}
%         \label{fig:subfig3}
%     \end{subfigure}
%     % \caption{Comparison of original and reconstructed speech spectrograms.}
%     \caption{Comparative analysis of speech spectrograms: original signal versus reconstructions from 2-token representations.}
%     \label{fig:threefigures}
% \end{figure*}
Speech codecs operate by converting encoded embedding features into tokens that correspond to the nearest values in a VQ codebook, followed by decoding the quantized features to reconstruct the original signal. 
\begin{table}[h]
\centering
\caption{Comparison of PESQ Scores for speech codec}
\label{tab:pesq_psnr_comparison}
\begin{tabular}{lccc}
\hline
Model & Bandwidth & token/s & PESQ $\uparrow$ \\
\hline
Encodec &  & 200 & 2.091 \\
SpeechTokenizer & 2.0kbps & 200 & 1.905 \\
Ours &  & 200 & 2.351 \\
\hline
Encodec &  & 400 & 2.621 \\
SpeechTokenizer & 4.0kbps & 400 & 2.597 \\
Ours &  & 400 & \textbf{3.085} \\
\hline
\end{tabular}
\end{table}
In our experimental evaluation, we compared our proposed method with two state-of-the-art approaches: Encodec and SpeechTokenizer. 
For quantitative assessment, we employed the Perceptual Evaluation of Speech Quality (PESQ) metric~\cite{PESQ}, which models the human auditory system to evaluate speech quality.
Table \ref{tab:pesq_psnr_comparison} shows the results obtained using the Librispeech test-clean dataset. We utilized a 10-bit codebook for token quantization and conducted experiments with frame rates of 200 and 400 tokens per second. 
Our model achieved a PESQ score of 2.351 with 4 tokens per frame, comparable to SpeechTokenizer using 8 tokens. With 8 tokens, our model outperformed baselines, reaching a PESQ score of 3.085.
Fig. \ref{fig:threefigures} illustrates the comparison between the original spectrogram and the spectrogram reconstructed using only two tokens. This demonstrates that our proposed approach effectively encapsulates semantic and acoustic information within the first and second tokens.

\subsection{Voice conversion}
To validate the representational capacity of our acoustic tokens, we conducted a voice conversion task, transforming source speech into target speech. For two speech signals, $x^{source}$ and $x^{target}$, we first converted each into embedding features through our Encoder, followed by quantization. 
We then performed a token swap operation, exchanging the second token—which encapsulates acoustic information—before reconstructing the signal through our decoder model. This process can be formalized as:
\begin{equation}
\mathbf{\hat{x}^{convert}} = D(\mathbf{\hat{z}}_0^{source}, \mathbf{\hat{z}}_1^{target})
\end{equation}
To evaluate the effectiveness of our approach, we compared our results with two relevant methods: FreeVC~\cite{freevc}, a state-of-the-art technique known for its effective feature extraction and swapping from target waveforms, and SpeechTokenizer. 
\begin{table}[h!]
\caption{Comparison of MCD and pitch correlation for voice conversion}
\label{tab:VC_results}
\centering
\begin{tabular}{c c c}
\hline
\textbf{}            & \textbf{MCD $\downarrow$} & \textbf{Pitch correlation $\uparrow$} \\ \hline
SpeechTokenizer      & 7.151                     & 0.001                                 \\ 
FreeVC               & 6.831                     & 0.239                                 \\ 
Ours                 & \textbf{6.076}                     & \textbf{0.310}                                 \\ \hline
\end{tabular}
\end{table}

\begin{table}[h!]
\caption{Speech Emotion Recognition (SER) accuracy for token-based models on the RAVDESS dataset}
\label{tab:SER_results}
\centering
\begin{tabular}{c c c}
\hline
\textbf{}            & token/s & Accuracy $\uparrow$ \\ \hline
Encodec             & 50                     & 0.500                                 \\ 
SpeechTokenizer     & 50                     & 0.455                                 \\ 
Ours                & 50                     & \textbf{0.553}                                 \\ \hline
\end{tabular}
\end{table}
Notably, FreeVC performs conversion directly on the waveform, while SpeechTokenizer, similar to our method, employs token swapping.
For quantitative evaluation, we employed two metrics: Mel Cepstral Distortion (MCD) between the converted speech and the target speech, and pitch correlation to assess the alignment of pitch characteristics. The results of these comparisons are presented in Table \ref{tab:VC_results}.
Our results show that the model's second token effectively captures acoustic information. When this token is swapped from target to source and decoded, our approach outperforms baseline methods in both MCD and pitch correlation metrics. This highlights our token-based representation's effectiveness in capturing and transferring acoustic characteristics, crucial for high-quality voice conversion. The improved performance in spectral (MCD) and prosodic (pitch correlation) domains indicates a more comprehensive transfer of voice characteristics compared to existing methods.

\subsection{Speech Emotion Recognition}
Speech signals inherently combine semantic content and acoustic features, including prosody. Prosodic elements play a crucial role in conveying emotional information. To assess the efficacy of our proposed method in capturing these nuanced aspects, we conducted experiments on SER.
For our evaluation, we utilized the Ryerson Audio-Visual Database of Emotional Speech and Song (RAVDESS), a widely recognized dataset in emotion recognition. We focused on 4 distinct emotion classes: neutral, happy, sad, and angry.
Our tokenization model was employed without any emotion-specific fine-tuning. For classification, we used a Bidirectional Encoder Representations from Transformers (BERT)~\cite{BERT2019}, augmented with a 4-class classifier layer. 
The BERT-based classifier was trained for 30 epochs and evaluated on RAVDESS speakers not included in the training set. We compared our method against two prominent baselines: Encodec and SpeechTokenizer.
Our experimental results indicate that the proposed model achieves superior emotion recognition accuracy compared to the baselines, as shown in Table \ref{tab:SER_results}. This success can be attributed to our model's ability to effectively capture and represent subtle prosodic variations critical for emotion detection in speech.

\subsection{Automatic Speech Recognition}
\begin{table}[h!]
\centering
\caption{Comparison of WER for ASR of multimodal LLM}
\label{tab:WER_for_ASR}
\begin{tabular}{c c c c}
\hline
\textbf{}     & LLM    & token/s & WER $\downarrow$ \\ \hline
SpeechTokenizer & \multirow{2}{*}{Qwen2-1.5B} & 50               & \textbf{0.253}        \\ %\cline{3-4} 
Ours            &                            & 50               & 0.290        \\ \hline
\end{tabular}
\end{table}
We performed experiments to evaluate our token-based multimodal LLM approach in comparison with existing methods, particularly focusing on the SpeechTokenizer technique. Our experiments utilized the Qwen2 model~\cite{qwen2} as the base LLM, trained for 3 epochs on the LibriSpeech 500-hour dataset to ensure consistency across compared methods.
The experimental setup employed the following prompt structure for the speech transcription task:
\begin{center}
\texttt{"Transcribe this speech"}
\end{center}
We evaluated ASR performance using the Word Error Rate (WER) metric on the LibriSpeech test-clean dataset. To facilitate effective learning between text and speech tokens within the Multimodal LLM framework, we adopted a token rate of 50 tokens per second, using a single token value per frame. This configuration was maintained consistently throughout both training and evaluation phases.
Our results showed that the Speech LLM trained with SpeechTokenizer achieved a WER of 0.253, slightly outperforming our proposed model, which attained a WER of 0.290. Table \ref{tab:WER_for_ASR} presents the detailed results of this evaluation.
A key observation is that despite the absence of explicit acoustic information in the SpeechTokenizer approach, our model achieved comparable performance. These findings suggest that while our proposed method may not surpass the SpeechTokenizer in terms of WER, it demonstrates competitive performance while retaining acoustic information. This characteristic could potentially offer advantages in tasks requiring fine-grained audio analysis.

\section{Conclusion}
\label{sec:print}
In this paper, we introduced a speech tokenizer modeling both semantic and acoustic information for various speech tasks. By incorporating acoustic guidance and loss, our method enhances feature representation, improving emotion recognition and speech signal understanding. Disentangling audio components allows for more efficient signal management. Future work could refine this approach with advanced models and apply it to multimodal LLMs for broader tasks, including prosody modeling. While promising, further research should explore its potential in multilingual processing and human-computer interaction, potentially transforming AI interactions in audio applications.

\bibliographystyle{IEEEbib}
\bibliography{refs}

\end{document}